\documentclass{article}
\usepackage{iclr2027_conference,times}
\usepackage[T1]{fontenc}
\usepackage{microtype}
\usepackage{graphicx,booktabs,tabularx,array,amsmath,amssymb}
\newcolumntype{L}[1]{>{\raggedright\arraybackslash}p{#1}}
\newcolumntype{Y}{>{\raggedright\arraybackslash}X}
\usepackage[font=small,labelfont=bf]{caption}
\usepackage{xcolor}
\usepackage{xurl}
\definecolor{linkblue}{RGB}{28,76,112}
\usepackage[colorlinks=true,allcolors=linkblue]{hyperref}

\usepackage{enumitem}
\setlist{leftmargin=*}
\newcommand{\dataset}{Spatial Foundry}

\newcommand{\wTc}{T^{\mathrm{wc}}}
\title{Spatial Foundry: A Multimodal Dataset\\for Spatial Intelligence}
\author{Ethan Lin, Yang Song
\\Ranolabs\\\href{mailto:support@ranolabs.com}{\texttt{support@ranolabs.com}}}
\date{}
\iclrfinalcopy
\begin{document}
\maketitle
\lhead{Spatial Foundry}
\begin{abstract}
Spatial intelligence requires connecting visual observations to the geometry, objects, and persistent state of an environment. Images alone leave these relationships implicit, while disconnected annotations make it difficult to study appearance, camera motion, and scene changes under a common reference. We introduce \dataset{}, a collection of 5,280 furnished residential 3D scene configurations with editable assets and aligned visual, geometric, and temporal data. Its modalities include floor plans, perspective and panoramic RGB, metric depth, point clouds, meshes, instance-level 3D bounding boxes, and camera trajectories paired with video. Shared scene coordinates link observations across views and time; editable source scenes enable additional camera paths and controlled object changes. We describe the data representation and construction workflow, and discuss applications in floor-plan-conditioned 3D generation, geometric reconstruction, instance-level perception, camera-conditioned video generation, shared world modeling, and scene editing. Controlled camera and scene variations support spatial perception and world-model research. The public project page provides videos and an interactive colored point-cloud preview.
\end{abstract}
\noindent\textbf{Project page:} \url{https://ranolabs.com/Blog/spatial-foundry}
\section{Introduction}
Understanding a three-dimensional environment requires more than recognizing objects in the current view. It requires knowing how objects are arranged in space and anticipating what would be observed from another viewpoint or after a change to the environment. Connecting visual observations to spatial structure and its changes is a central aspect of spatial intelligence. World models offer a way to learn these relationships by predicting subsequent observations conditioned on prior observations, camera motion, and actions that modify the scene. In three-dimensional environments, these predictions should be visually plausible, follow the specified camera trajectory, and preserve consistent geometry and object states across viewpoints and revisits. Learning and evaluating these capabilities requires images and videos explicitly associated with the underlying geometry, camera poses, and scene states, so that visual changes can be traced to their spatial causes.

Existing datasets and simulation environments provide complementary foundations for spatial learning. RealEstate10K associates video observations with recovered camera trajectories~\citep{realestate}; Hypersim pairs photorealistic indoor imagery with geometric and semantic annotations~\citep{hypersim}; and ProcTHOR provides diverse procedural environments for embodied interaction~\citep{procthor}. As summarized in Figure~\ref{fig:datacomparison}, these resources differ in their coverage of paired layouts, visual observations, geometry, and editable scene states. Access to individual modalities does not by itself provide matched sequences in which camera motion and object changes can be varied independently. Studying such controlled spatial variation requires observations, camera trajectories, and persistent object identities coordinated within a common scene reference: the same route observed before and after an object edit, or the same scene state observed from multiple viewpoints.

\begin{figure*}[t]

\centering
\begingroup
\definecolor{coveragegreen}{RGB}{24,117,101}
\definecolor{coveragegold}{RGB}{169,123,37}
\newcommand{\covD}{\textcolor{coveragegreen}{$\bullet$}}
\newcommand{\covR}{\textcolor{coveragegold}{$\triangle$}}
\newcommand{\covN}{\textcolor{gray}{--}}
\newcommand{\covhead}[1]{\shortstack[c]{#1}}
\setlength{\tabcolsep}{1.1pt}
\renewcommand{\arraystretch}{1.5}
\scriptsize
\begin{tabularx}{\textwidth}{@{}p{.225\textwidth}*{10}{>{\centering\arraybackslash}X}@{}}
\toprule
 & Layout & \multicolumn{4}{c}{Visual observations} & \multicolumn{3}{c}{Geometry} & \multicolumn{2}{c}{Control}\\
\cmidrule(lr){2-2}\cmidrule(lr){3-6}\cmidrule(lr){7-9}\cmidrule(l){10-11}
Dataset & \covhead{Plan--3D\\pairs} & \covhead{Persp.\\RGB} & \covhead{360$^\circ$\\RGB} & \covhead{Video /\\poses} & \covhead{Metric\\depth} & \covhead{3D\\geometry} & \covhead{Point\\cloud} & \covhead{Boxes /\\instances} & \covhead{Editable\\assets} & \covhead{State\\edits}\\
\midrule
RealEstate10K\newline{\tiny\citep{realestate}} & \covN & \covD & \covN & \covD & \covN & \covN & \covN & \covN & \covN & \covN\\
ScanNet++\newline{\tiny\citep{scannetpp}} & \covR & \covD & \covD & \covD & \covD & \covD & \covD & \covR & \covN & \covN\\
Structured3D\newline{\tiny\citep{structured3d}} & \covD & \covD & \covD & \covN & \covD & \covD & \covR & \covD & \covN & \covN\\
3D-FRONT\newline{\tiny\citep{front}} & \covD & \covR & \covR & \covR & \covR & \covD & \covR & \covR & \covD & \covR\\
Hypersim\newline{\tiny\citep{hypersim}} & \covR & \covD & \covN & \covD & \covD & \covR & \covR & \covD & \covR & \covR\\
ProcTHOR\newline{\tiny\citep{procthor}} & \covD & \covR & \covR & \covR & \covR & \covD & \covR & \covR & \covD & \covR\\
\midrule
\textbf{Spatial Foundry}\newline{\tiny 5,280 configurations} & \covD & \covD & \covD & \covD & \covD & \covD & \covD & \covD & \covD & \covR\\
\bottomrule
\end{tabularx}
\par\vspace{4pt}
\begin{minipage}{\textwidth}\scriptsize
\covD\ Documented released product\quad
\covR\ Generated / derived; source access may be required\quad
\covN\ Not specified in the compared release
\end{minipage}
\endgroup

\caption{\textbf{Data modalities and scene controllability.} Representative resources span captured video~\citep{realestate}, indoor scans~\citep{scannetpp}, synthetic structure and appearance~\citep{structured3d,front,hypersim}, and interactive environments~\citep{procthor}. Filled circles identify documented released products; triangles identify products obtained through rendering, processing, or source-asset access. Dashes indicate features not specified in the compared release. Availability is described at resource level. Spatial Foundry combines 5,280 furnished configurations with aligned observations and editable source scenes.}
\label{fig:datacomparison}
\end{figure*}

To support these controlled comparisons, we introduce \dataset{}, a collection of 5,280 furnished residential 3D scene configurations and a scene-based pipeline for producing aligned visual, geometric, and temporal data. Each configuration combines a residential layout with a furnishing arrangement. GPT-6 generates structured layouts from residential rules, human reviewers check their spatial usability, and Blender assembly with Infinigen assets produces the furnished source scenes. The collection spans US-style residential layouts with varied room compositions and furnishing arrangements. Editable source scenes retain metric geometry and object structure, supporting perspective and panoramic RGB observations, depth maps, meshes, point clouds, and instance-level 3D bounding boxes. Camera trajectories associate video observations with per-frame camera poses, while scene edits provide explicit records of changes to object placement and appearance. The source scenes allow additional observations to be rendered under specified cameras and object configurations.

The central organizing principle is to separate \emph{changes in observation} from \emph{changes in the environment}. With a scene held fixed, different camera trajectories produce observations of the same geometry and object configuration. With a camera trajectory held fixed, controlled edits produce corresponding sequences before and after a scene change. Multiple observer cameras can additionally record a shared sequence of scene states. These controlled comparisons connect appearance to the spatial variables that determine it, supporting supervision and evaluation beyond isolated images. Potential applications include geometric reconstruction, 3D instance perception, camera-conditioned video generation, and prediction of the visual consequences of scene edits.

This data organization supports floor-plan-conditioned 3D scene generation, geometric reconstruction, instance-level spatial perception, camera-conditioned video generation, multi-agent shared world modeling, and state-conditioned scene editing. Paired floor plans and furnished environments connect layout constraints to 3D structure and appearance, while aligned observations and state records connect camera motion and object changes to their visual consequences. Together, these applications span static scene understanding and the prediction of observations during exploration and interaction.

This report makes three contributions:
\begin{enumerate}
\item A collection of furnished residential environments with metric geometry, editable assets, and diverse layout and furnishing configurations.
\item A pipeline for aligned multimodal and sequential data, linking rendered observations to camera calibration, trajectories, geometry, object annotations, and scene-state records.
\item Controlled data-generation workflows that separate camera variation from scene variation, supporting reproducible comparisons for spatial perception and generation.
\end{enumerate}

\begin{figure*}[t]
\centering
\includegraphics[width=\textwidth]{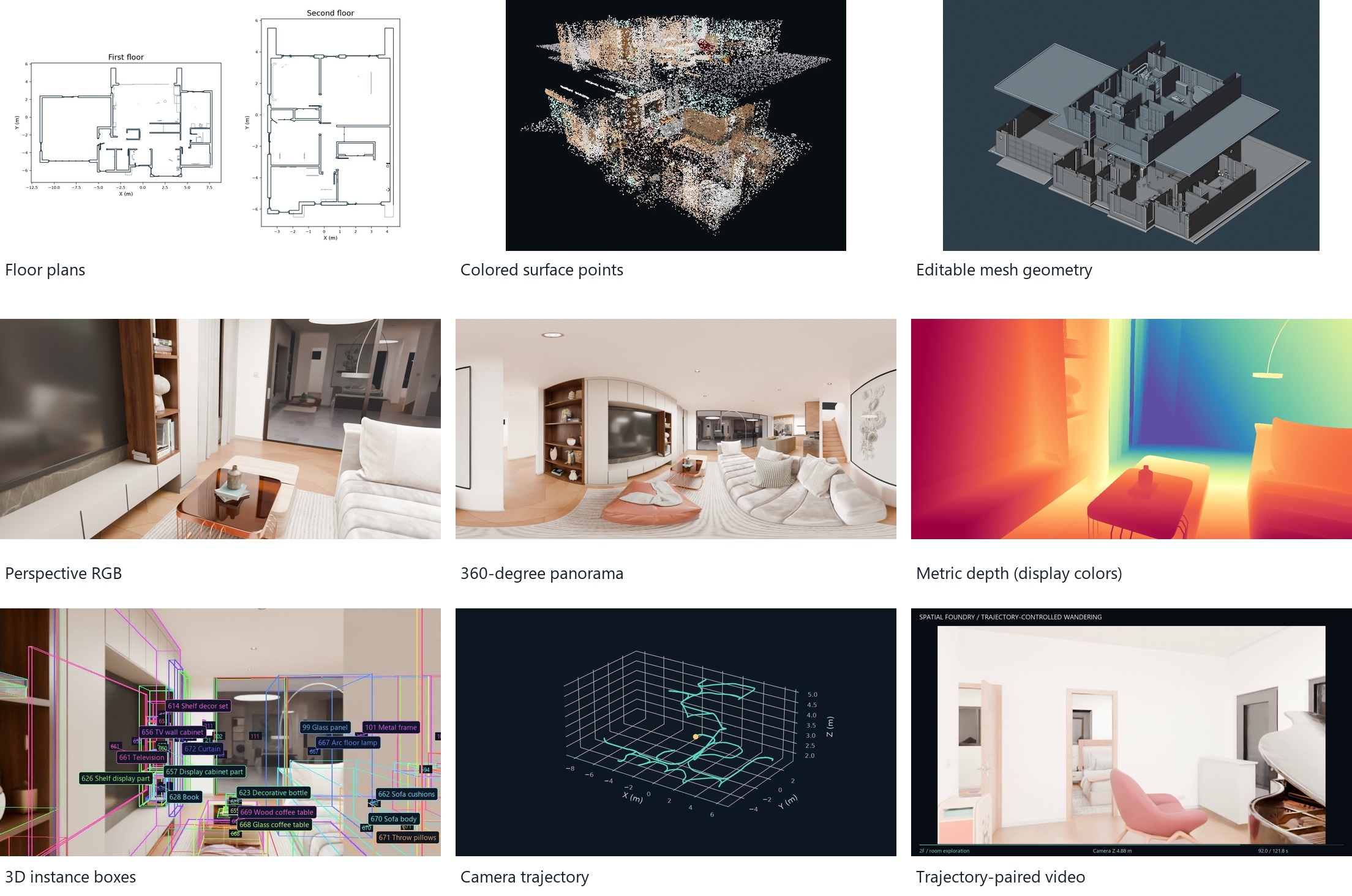}
\caption{\textbf{Aligned observations and 3D representations.} The two-floor scene provides floor plans, a colored surface-point preview, editable mesh geometry, perspective RGB, a panorama, metric-depth visualization, instance-box overlays, and a camera trajectory. This static overview uses UV textures and material Base Color for point coloring, display-space floor separation for interior visibility, and a color map for depth visualization.}
\label{fig:overview}
\end{figure*}

\section{Related Work}
\subsection{Indoor 3D Datasets and Simulation Environments}
Captured indoor datasets connect visual observations with reconstructed geometry. ScanNet++ provides high-fidelity scans, registered imagery, and semantic annotations~\citep{scannetpp}, while Replica and HM3D supply reconstructed indoor environments for visual and embodied research~\citep{replica,hm3d}. Synthetic resources expose additional structure: Structured3D pairs rendered observations with structural annotations~\citep{structured3d}, 3D-FRONT provides furnished layouts and textured assets~\citep{front}, and Hypersim associates photorealistic imagery with dense geometric and semantic supervision~\citep{hypersim}. ProcTHOR supports embodied interaction in procedurally generated environments~\citep{procthor}, while Aria Synthetic Environments provides synthetic egocentric observations with camera trajectories and structured scene annotations~\citep{ase}.

These resources emphasize captured appearance, structural understanding, furnished assets, or interactive simulation. For studying controlled spatial changes, recorded views and fixed observation sequences alone provide limited access to matched camera and object variations: observing the same route before and after a chosen edit requires editable geometry and coordinated state records. Spatial Foundry organizes 5,280 furnished residential configurations around paired 2D floor plans and 3D environments, with perspective and panoramic RGB, metric depth, meshes, point clouds, instance annotations, and trajectory--video correspondences. Its scope connects layout-conditioned generation and static perception to sequential observation within the same residential assets. Editable scenes allow camera paths and object configurations to be varied independently, yielding task-specific pairs with explicit geometric and state references.

\subsection{Controllable Generation and Spatial World Models}
Camera-conditioned generation and interactive world models aim to predict observations under specified motion or actions. RealEstate10K supplies video frames with camera trajectories recovered through SLAM and bundle adjustment~\citep{realestate}. CameraCtrl introduces explicit camera-pose conditioning for video diffusion, and CameraCtrl II extends camera-controlled generation to dynamic scene exploration~\citep{cameractrl,cameractrlii}. Recurrent world models learn compact predictive environment representations~\citep{worldmodels}; Genie learns action-controllable environments from unlabeled video through a latent action model~\citep{genie}. Atlas combines visual and geometric inputs in a shared spatial context for camera-controlled generation, reconstruction, and simulation~\citep{atlas}. VSI-Bench evaluates visual-spatial reasoning from sequential video observations~\citep{vsibench}.

These approaches make camera control, persistent context, and interaction central to visual generation. Their supervision also highlights a data requirement: recovered camera trajectories describe viewpoint changes, whereas latent actions learned from video do not directly specify metric object transforms. Relating a visual change to a particular camera motion or object edit requires corresponding geometric and state records. Spatial Foundry supplies these references through calibrated observations, paired camera trajectories and videos, persistent instance identities, and recorded scene edits. Holding the scene fixed provides camera-conditioned sequences; holding the camera path fixed provides before-and-after edit pairs. Multiple cameras can observe the same state sequence, extending this supervision to cross-view consistency, revisitation, and shared world modeling.

\section{Dataset Overview}
\subsection{Residential Environments}
The collection contains 5,280 furnished residential 3D scene configurations. Each configuration combines an architectural layout, structural geometry, and a particular arrangement of furniture and objects. The scenes are generated from U.S. residential design priors, GPT-6 floor-plan definitions, and Infinigen-based Blender assembly. Human review checks room connectivity, passage clearance, and stair access. Variation in room combinations, circulation paths, functional templates, and furnishing arrangements produces diverse US-style residential environments. Multiple furnishing configurations can share a layout, providing variation in both architectural structure and interior contents.

The source assets retain explicit geometry and object transforms. Distances and object dimensions use meters. Floor plans describe the arrangement of rooms and structural boundaries; meshes represent surfaces; point clouds provide sampled geometry. The stored geometry retains its original metric coordinates.

\subsection{Modalities and Correspondence}
\label{sec:modalities}
For scene $s$, let $\mathcal{G}_s$ denote geometry and $\mathcal{O}_s$ the object-instance table. Observation $t$ is represented conceptually as
\begin{equation}
o_{s,t}=\big(I_{s,t},D_{s,t},\Pi_{s,t},\wTc_{s,t},\tau_{s,t}\big),
\end{equation}
where $I$ is RGB, $D$ is metric depth, $\Pi$ specifies the projection model and calibration, $\wTc$ maps camera coordinates to world coordinates, and $\tau$ is time. Instance references connect observations to $\mathcal{O}_s$. This representation expresses the shared spatial and temporal associations among the modalities.

Perspective observations use calibrated pinhole views. Panoramic observations use an equirectangular camera model with a declared orientation. Numeric depth is stored with its units and valid-pixel mask, alongside colorized previews for visualization. Object records include instance identifiers, names or semantic descriptions where available, metric bounds, and transforms. Persistent instance identifiers establish correspondence across views within a scene, while semantic descriptions characterize the associated objects.

Trajectory--video pairs associate each native frame with a camera pose and timestamp. Frame-level records provide the camera supervision for temporal tasks. Presentation videos retain source-frame associations when playback timing is adjusted. The application demonstrations illustrate these correspondences along room-to-room and multi-floor routes.

\begin{table}[t]
\centering\small
\begin{tabularx}{\columnwidth}{@{}L{0.31\columnwidth}Y@{}}
\toprule
Modality & Spatial or temporal reference \\
\midrule
Floor plans & Layout, room boundaries, metric geometry \\
Perspective / 360 RGB & Camera model, pose, scene identity \\
Metric depth & Projection, units, depth convention, mask \\
Point clouds / meshes & World coordinates, scale, object association \\
3D instances & ID, bounds, dimensions, transform, label provenance \\
Trajectories / video & Native frame ID, time, camera pose, intrinsics \\
Scene edits & Event, affected instance, before/after state \\
\bottomrule
\end{tabularx}
\caption{\textbf{Aligned data modalities.} Coordinates, camera calibration, instance identifiers, and timestamps connect the representations within each scene.}
\label{tab:modalities}
\end{table}

\subsection{Interactive Visualization}
The public project page provides videos, images, and an interactive colored point-cloud preview. Other demonstration clouds use texture/material or rendered-RGB color estimates, with the color source recorded for each product. These lightweight previews support interactive inspection of the scene geometry and complement the full source scenes and numeric data products.

\section{Scene Construction and Multimodal Data Generation}
\label{sec:construction}
\begin{figure*}[t]

\centering\includegraphics[width=\textwidth]{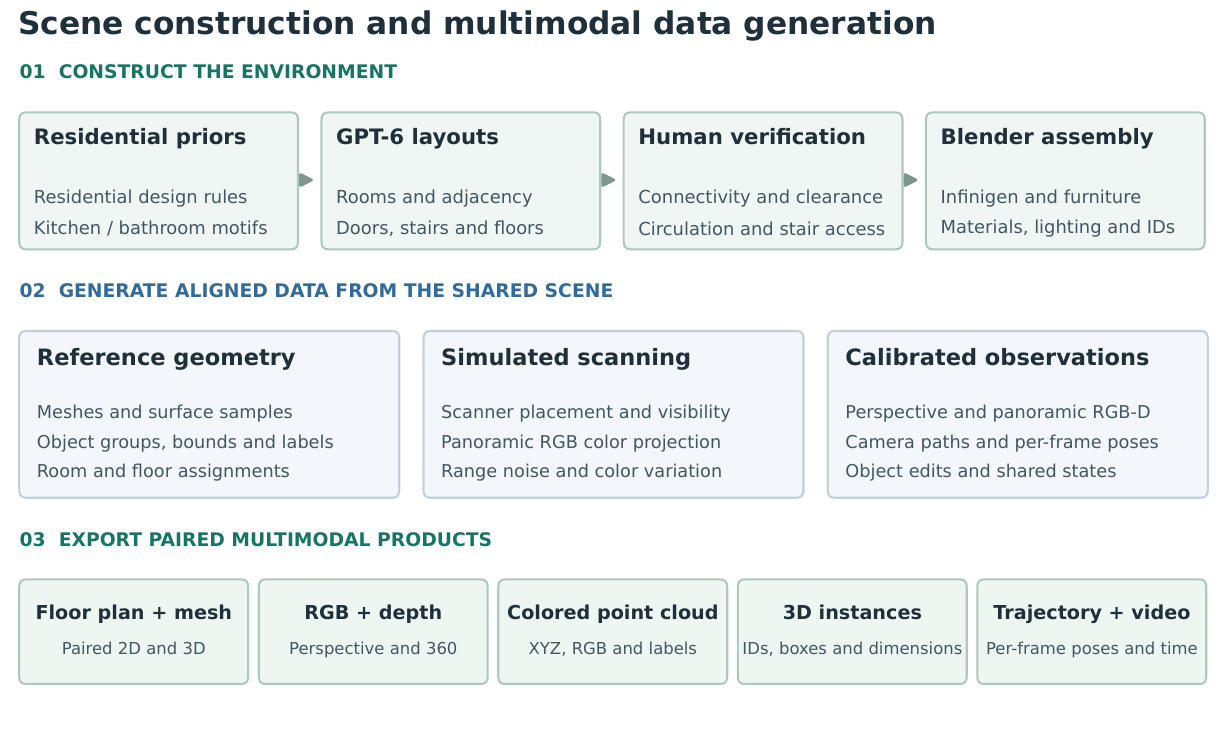}
\caption{\textbf{Scene construction and data generation.} Residential rules and functional motifs condition GPT-6 layout generation. Human review checks doorways, room circulation, and stair connectivity. Blender assembly combines the accepted layout with Infinigen assets, materials, and furniture. Reference geometry supports virtual scanning and calibrated rendering, while camera trajectories and object-state changes produce aligned temporal observations.}
\label{fig:construction}
\end{figure*}

The construction workflow connects residential design priors to editable Blender scenes and aligned synthetic observations (Figure~\ref{fig:construction}). GPT-6 generates structured floor-plan definitions from residential rules and functional layout templates. Human review checks spatial organization and circulation before scene assembly. Infinigen and a reusable asset and material library provide the architectural and furnishing content. The completed scene then supplies a common reference for point-cloud simulation, RGB-D rendering, instance annotations, and controlled temporal observations.

\subsection{Rule-Conditioned Layout Generation and Human Review}
\begin{figure*}[t]

\centering\includegraphics[width=.94\textwidth]{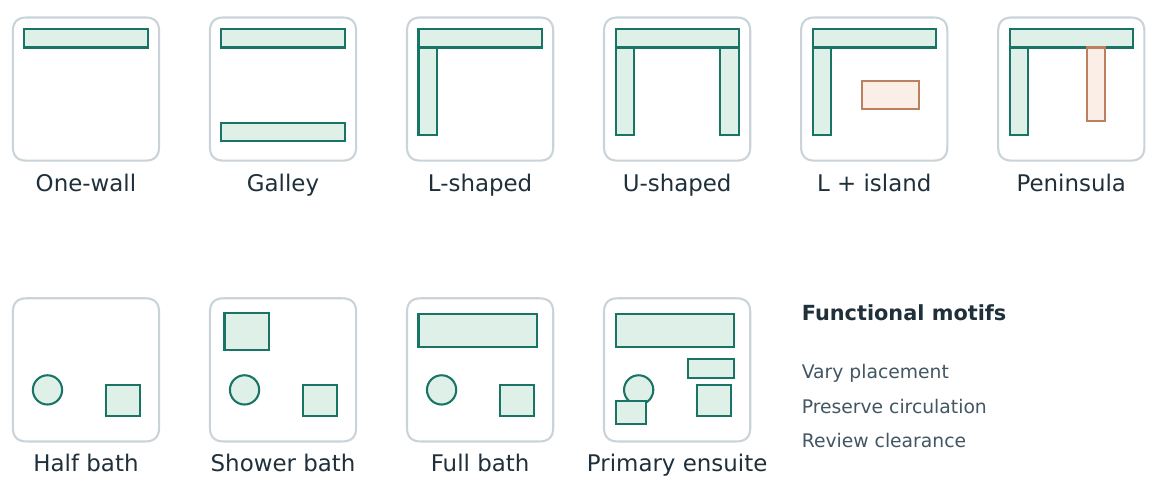}
\caption{\textbf{Functional layout motifs.} Kitchen arrangements vary from one-wall and galley layouts to L-shaped, U-shaped, island, and peninsula configurations. Bathroom motifs vary the combination and placement of fixtures. These schematic patterns illustrate the functional priors used to diversify layouts and guide circulation review.}
\label{fig:motifs}
\end{figure*}

U.S. residential design guidance, including the International Residential Code~\citep{irc}, provides priors for openings, circulation, and stair organization. We encode these requirements together with construction parameters, such as wall thickness, into a structured rule set. The design reference organizes these priors at three spatial scales: dwelling footprint and floor organization, room-level functional layouts, and fixture and storage placement. Footprint templates include compact rectilinear plans and winged arrangements such as L-, T-, and U-shaped plans; floor organization specifies how rooms and circulation connect across levels. At room scale, kitchen templates vary counter arrangements and island or peninsula placement, while bathroom templates vary fixture combinations and access from a hall, an adjacent bedroom, or a primary suite (Figure~\ref{fig:motifs}). These categories provide a structured vocabulary for layout generation, connecting the US-style character of the scenes to architectural and functional organization.

Let $\mathcal{R}$ denote the rule set, $b$ the requested room and floor program, and $m$ a functional motif. GPT-6 produces a structured layout $P$ conditioned on these inputs:
\begin{equation}
P\sim q_{\mathrm{GPT\text{-}6}}(P\mid b,\mathcal{R},m),\qquad
P=\big(\{\Omega_r\},\mathcal{W},\mathcal{D},\mathcal{S},\mathcal{A}\big),
\label{eq:layoutproposal}
\end{equation}
where $\Omega_r$ are room polygons, $\mathcal{W}$ the wall specifications, $\mathcal{D}$ the openings, $\mathcal{S}$ the stairs and floor connections, and $\mathcal{A}$ the room adjacency relations. This structured definition links the 2D plan to the geometry subsequently instantiated in Blender.

Functional relationships further guide layout proposals and review. Storage is positioned in relation to the spaces it serves: entry storage near the entrance, bedroom closets near sleeping areas, linen storage near bathrooms, and pantry storage near the kitchen. Kitchen work areas and bathroom fixtures are arranged with surrounding circulation space, while the grouping of kitchens, bathrooms, and laundry areas informs service-zone adjacency. For multi-floor layouts, stair placement connects circulation areas and aligns stairwells, floor openings, and landings across levels.

Human reviewers inspect candidate plans and their spatial realization for room connectivity, doorway clearance, blocked passages, and usable stair access between floors. Layouts are revised or filtered according to these checks. Review is repeated after furnishing to identify objects that obstruct doors, circulation, or stair landings.

\subsection{Blender Assembly and Furnishing}
Accepted plans are instantiated as metric Blender scenes with floors, walls, openings, stairs, and room-level assignments. A reusable rendering library supplies surface materials such as wall finishes, floor coverings, and tiles. Infinigen~\citep{infinigenindoors} and GPT-6-assisted procedural generation provide assets and scene-construction instructions for furnishing the layout. Object placement follows the room function and reviewed circulation constraints.

We represent the assembled scene as
\begin{equation}
S=\big(\mathcal{G},\mathcal{O},\mathcal{M},\mathcal{L}\big),\qquad
o_i=\big(\iota_i,\mathcal{G}_i,T_i,\ell_i\big),
\label{eq:scenestate}
\end{equation}
where $\mathcal{G}$ is geometry, $\mathcal{O}$ the instance table, $\mathcal{M}$ the materials, and $\mathcal{L}$ the lighting. Each instance associates an identifier $\iota_i$ with its geometry $\mathcal{G}_i$, transform $T_i$, and semantic description $\ell_i$. Scene construction preserves the correspondence between plan elements and their 3D realization, producing paired floor plans and furnished environments. Changing the furnishing or material choices while retaining the layout yields additional configurations under the same structural constraints.

\subsection{Reference Geometry and Instance Annotations}
Post-processing extracts evaluated mesh geometry in a common world coordinate system. Geometric grouping, semantic inference, and review associate component meshes with object instances. A reference surface cloud is sampled from the scene, retaining the instance and semantic associations of its source geometry. This reference is kept separately from the simulated scan, so observation perturbations preserve the underlying annotation coordinates.

For world-space vertices $\{\mathbf{v}_{i,j}\}$ assigned to instance $i$, its axis-aligned bounds and dimensions are
\begin{equation}
\mathbf{b}^{-}_i=\min_j\mathbf{v}_{i,j},\qquad
\mathbf{b}^{+}_i=\max_j\mathbf{v}_{i,j},\qquad
\mathbf{d}_i=\mathbf{b}^{+}_i-\mathbf{b}^{-}_i,
\label{eq:instancebounds}
\end{equation}
with extrema taken componentwise. Architectural geometry also supplies walls, openings, room boundaries, and floor assignments. These records connect object annotations, point labels, rendered views, and paired 2D layouts.

\subsection{Virtual Scanner Placement and Visibility}
The mobile-scan simulation combines reference surface samples, virtual scanner locations, rendered appearance, and geometric perturbations. Scanner positions are distributed within rooms using clearance and coverage checks. Each scanner has a known position $\mathbf{s}_k$ and orientation $R_k$, and a corresponding RGB panorama is rendered at that pose.

For reference point $\mathbf{p}_j$, its scanner-frame direction and range are
\begin{equation}
\mathbf{q}_{k,j}=R_k^{\top}(\mathbf{p}_j-\mathbf{s}_k),\qquad
r_{k,j}=\|\mathbf{q}_{k,j}\|_2.
\end{equation}
Spherical projection assigns the direction to an angular bin $b_{k,j}$. Visibility is approximated by retaining the nearest sampled surface point in each bin:
\begin{equation}
\mathcal{V}_k=\left\{j:\;j=\arg\min_{i:\,b_{k,i}=b_{k,j}}r_{k,i}\right\},\qquad
\mathcal{V}=\bigcup_k\mathcal{V}_k.
\label{eq:visibility}
\end{equation}
This operation approximates scanner occlusion while combining coverage from multiple room locations. Each retained point is assigned to the nearest scanner from which it is visible:
\begin{equation}
k^{\star}(j)=\arg\min_{k:\,j\in\mathcal{V}_k}r_{k,j}.
\label{eq:scannerassignment}
\end{equation}
The assignment supplies a shared observation origin for point visibility, rendered color, and range-dependent perturbation.

\subsection{Panoramic Color and Scan Perturbations}
Let $\mathbf{q}=(q_x,q_y,q_z)^\top$ be a point direction in an OpenCV-style scanner frame, with $x$ right, $y$ down, and $z$ forward. Its panorama coordinates are
\begin{equation}
\begin{aligned}
\theta&=\operatorname{atan2}(q_x,q_z),\qquad
\phi=\operatorname{atan2}\!\left(q_y,\sqrt{q_x^2+q_z^2}\right),\\
(u,v)&=\left(W\left(\frac{\theta}{2\pi}+\frac12\right),H\left(\frac{\phi}{\pi}+\frac12\right)\right).
\end{aligned}
\label{eq:panocolor}
\end{equation}
The color of point $j$ is sampled from the panorama $I_{k^{\star}(j)}$ at these coordinates. Small scanner-level exposure and white-balance changes introduce appearance variation while preserving the point-to-observation association.

Distance-dependent geometric noise and a mixture of larger perturbations introduce selected characteristics of mobile scans. We express the simulated observation as
\begin{equation}
\widetilde{\mathbf{p}}_j=\mathbf{p}_j+\boldsymbol{\epsilon}_j,\qquad
\boldsymbol{\epsilon}_j\sim(1-\eta)\mathcal{D}_{\mathrm{local}}(r_j)+\eta\mathcal{D}_{\mathrm{outlier}}(r_j),
\label{eq:scannoise}
\end{equation}
where $r_j=r_{k^{\star}(j),j}$, $\eta$ is the outlier mixing probability, $\mathcal{D}_{\mathrm{local}}$ describes the range-dependent perturbation, and $\mathcal{D}_{\mathrm{outlier}}$ the larger-error component. Reference geometry and instance bounds remain attached to the original scene. A separate ray-casting variant models selected transmission, reflection, and missing-return effects around glass and mirrors.

\subsection{Calibrated RGB-D and Temporal Observations}
The same scene generates perspective RGB, panoramas, and depth using saved camera poses and intrinsics. For camera-to-world transform $\wTc=[R\mid\mathbf{t}]$, intrinsic matrix $K$, and camera-axis depth $z$ at pixel $\widetilde{\mathbf{u}}=(u,v,1)^\top$, the corresponding point is
\begin{equation}
\mathbf{x}_w=R\left(zK^{-1}\widetilde{\mathbf{u}}\right)+\mathbf{t}.
\label{eq:unproject}
\end{equation}
Radial range uses the normalized camera ray instead. The exported camera record specifies the depth convention and coordinate transform. A world-space box corner is projected with $K R^{\top}(\mathbf{x}_w-\mathbf{t})$, followed by homogeneous normalization, connecting instance annotations to the same RGB-D observation.

Camera trajectories connect rooms and floors through positions $\mathbf{p}(t)$ and orientations $R(t)$. Height follows the local supporting floor and changes continuously on stairs. Speed and angular velocity vary to capture room structure and object details along the route. Native frames retain timestamps, camera calibration, and poses; outputs are organized at scene, room, and floor levels.

\subsection{Controlled Edits and Shared Observations}
Object translation, rotation, insertion, removal, and appearance changes update explicit scene state. For action $a_t$ and observer camera $C_{t+1}^{(k)}$,
\begin{equation}
S_{t+1}=F(S_t,a_t),\qquad
I_{t+1}^{(k)}=\mathcal{R}_{\mathrm{render}}\!\left(S_{t+1},C_{t+1}^{(k)}\right).
\end{equation}
Fixed camera paths yield corresponding observations before and after an edit. Multiple observer cameras share the same object-state sequence, providing cross-view observations of each change. Persistent instance identifiers and edit records connect these temporal observations to the reference geometry and object annotations.

\section{Applications}
\dataset{} connects floor plans, furnished 3D scenes, visual observations, camera trajectories, and object states. These correspondences support six application areas, from generating environments from spatial layouts to predicting observations under camera motion and scene edits. Editable source assets allow task-specific observations and controlled variations to be produced within the same spatial reference.

\subsection{Floor-Plan-Conditioned 3D Scene Generation}
Each residential scene associates a 2D floor plan with its corresponding furnished 3D environment. Across 5,280 scene configurations, these pairs cover variation in architectural layout, room composition, circulation, and furniture arrangement. They provide large-scale paired data for learning the relationship between a compact spatial layout and a complete three-dimensional interior. Multiple furnishing configurations associated with a layout further expose variation in interior contents under shared structural constraints.

Given a floor plan, a model can predict room geometry, recover structural organization, and arrange furniture within the resulting space. The corresponding 3D scene supplies reference surfaces, object placements, and metric dimensions, while rendered observations provide complementary appearance supervision. This pairing supports floor-plan-to-3D generation, furnished scene completion, layout-conditioned interior synthesis, and retrieval of 3D environments from 2D plans. The diversity of paired environments connects these tasks to a broad range of residential configurations.

\subsection{Geometric Reconstruction and Multimodal Scene Representation}
Perspective and panoramic RGB observations are associated with camera calibration, metric depth, and explicit scene geometry. Given one or more images, a model can estimate depth, recover three-dimensional structure, or construct a representation that supports novel views. Meshes and point clouds provide geometric references, while corresponding rendered views connect recovered structure to visual appearance.

Shared coordinates and camera models support calibrated reprojection, RGB-D fusion, and correspondence across perspective and panoramic observations. The editable scenes also allow additional camera views to be rendered for a given geometric configuration. These data support monocular and panoramic depth estimation, multiview reconstruction, novel-view synthesis, and the learning of neural scene representations with metric spatial supervision.

\subsection{Instance-Level Spatial Perception}
Object-instance records associate scene geometry with identifiers, transforms, bounding boxes, dimensions, and available semantic descriptions. Linking these records to RGB, depth, and point clouds supports tasks that connect visual evidence to individual objects in three-dimensional space. Given an observation, a model can localize an object, estimate its extent, and associate it with observations from other viewpoints.

Metric geometry provides references for object size, relative position, distance, and visibility. Persistent instance identifiers support cross-view association, while component-to-object groupings connect detailed meshes to furniture-level entities. These associations support 3D object detection, object grounding, spatial relation prediction, and object-centric scene representations. They also provide a basis for constructing spatial queries whose answers are derived from scene geometry and instance records.

\subsection{Camera-Conditioned Video Generation and Spatial Persistence}
Trajectory--video pairs associate visual observations with explicit camera positions, orientations, and timestamps. Given reference observations and a target camera path, a model can learn to generate the corresponding sequence of views. Position, viewing direction, elevation, and motion timing provide distinct controls over how the environment is observed.

Connected rooms and multi-floor routes extend this task beyond local viewpoint changes. Long trajectories expose new regions, introduce occlusions, and return to previously observed locations. The corresponding geometry and instance identities provide references for maintaining room layout, object placement, and cross-view consistency throughout the sequence. This setting supports camera-conditioned video generation, spatially grounded world modeling, and persistent scene representations under exploration and revisitation.

\subsection{Multi-Agent Collaboration and Shared World Modeling}
Multiple observer cameras can be associated with a common sequence of scene states. Each observer receives a partial view determined by its position, orientation, and occlusions, while object identities and scene changes remain shared. This organization provides aligned observations for learning how different agents perceive and update a common environment.

Given an action and observations from one or more viewpoints, a model can predict its visual consequences for other observers and update a shared spatial representation. Object-state records specify the affected instances and their changes, while camera geometry determines where those changes become visible. The resulting data support multi-agent observation prediction, cross-view state association, collaborative scene understanding, and online world modeling in which observations and edits update a persistent shared state.

\subsection{State-Conditioned Scene Editing}
Editable source scenes support controlled object translation, rotation, insertion, removal, and appearance changes. Applying these operations to the 3D scene produces corresponding observations with updated geometry, occlusion, and object relationships. Recording the edit together with the before-and-after states connects each operation to its spatial and visual consequences.

Holding the camera path fixed yields paired sequences that isolate scene changes; varying the viewpoints reveals the same edited state from different locations. These controlled pairs support action-conditioned observation generation, edit localization, view-consistent scene editing, and prediction of persistent object changes. The source assets also enable task-specific data augmentation through changes to object configuration, materials, and lighting, with camera and scene controls recorded separately.

\begin{figure*}[t]
\centering\includegraphics[width=\textwidth]{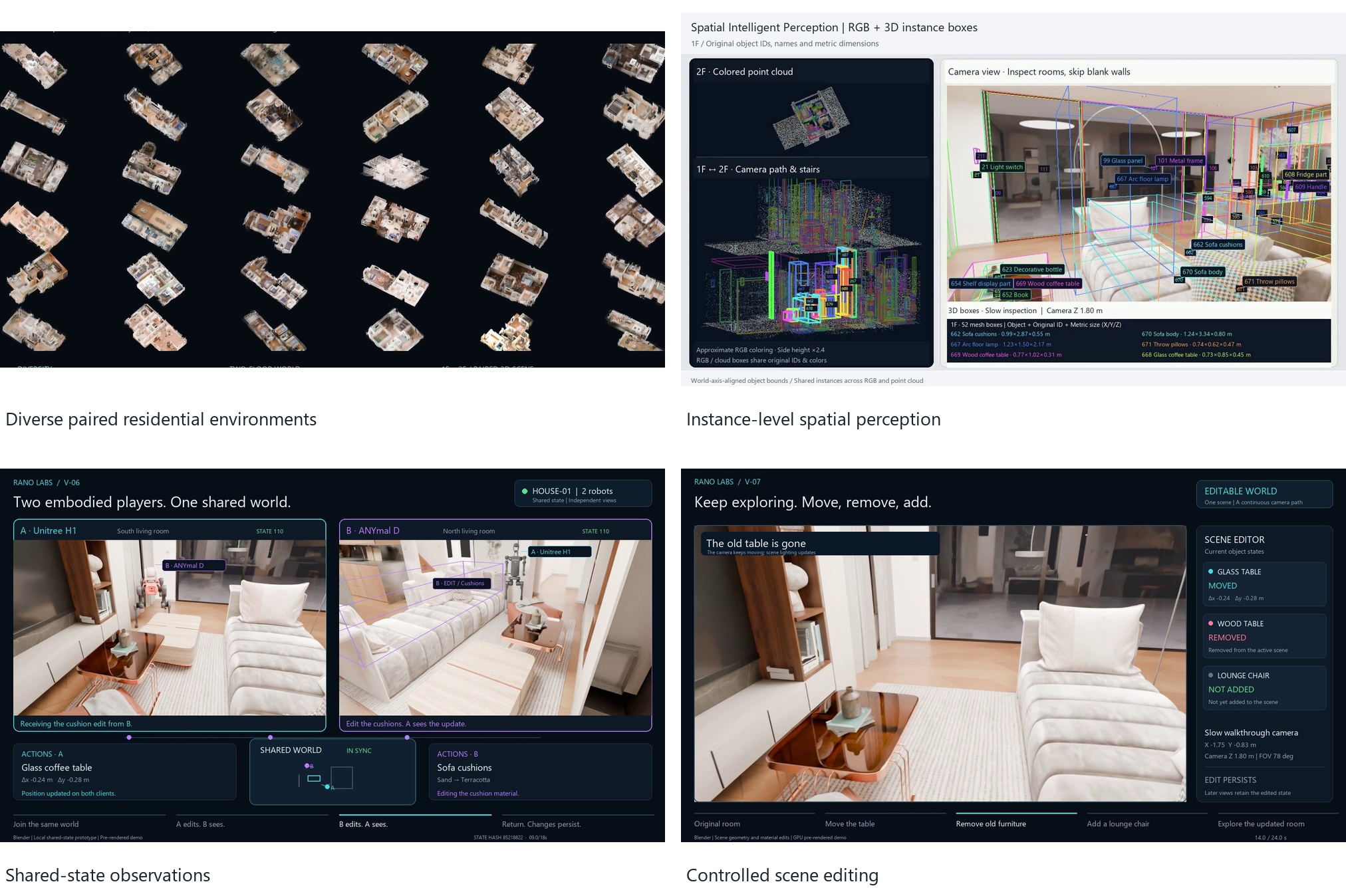}
\caption{\textbf{Application areas supported by aligned scene data.} Diverse furnished environments provide paired layout and 3D content. Instance annotations connect observations to objects and metric geometry. Shared scene states support multiple observer viewpoints, while editable assets provide controlled object changes. The examples illustrate the data correspondences used for spatial perception, generation, and world modeling.}
\label{fig:applications}
\end{figure*}

\bibliographystyle{iclr2027_conference}
\bibliography{references}
\end{document}